\documentclass{article}
\pdfoutput=1

\usepackage[nonatbib,preprint]{neurips_2019}

\usepackage[utf8]{inputenc} 
\usepackage[T1]{fontenc}    
\usepackage{hyperref}       
\usepackage{url}            
\usepackage{booktabs}       
\usepackage{amsfonts}       
\usepackage{nicefrac}       
\usepackage{microtype}      
\usepackage[numbers]{natbib}

\usepackage{wrapfig}

\usepackage{qiangstyle}
\usepackage{amsmath, amsfonts}
\usepackage{graphicx}
\usepackage{algorithm}
\usepackage{algorithmic}
\usepackage{float}
\usepackage[font=small]{caption}
\usepackage[compact]{titlesec}
\newcommand{\E}{\mathbb{E}}

\usepackage{xcolor}
\title{Sequence Modeling of Temporal Credit Assignment for Episodic Reinforcement Learning}

\author{%
Yang Liu \\ UIUC \\
\And 
Yunan Luo \\ UIUC\\
\And
Yuanyi Zhong \\ UIUC\\
\And
Xi Chen \\ covariant.ai \\
\And
Qiang Liu \\ UT Austin \\ 
\And
Jian Peng \\ UIUC \\
}

\newcommand{\edit}[1] {{\color{black}{#1}}}
\newcommand{\replace}[2] {{\color{black}{#1}}}

\begin{document}

\maketitle
\begin{abstract}
Recent advances in deep reinforcement learning algorithms have shown great potential and success for solving many challenging real-world problems, including Go game and robotic applications. Usually, these algorithms need a carefully designed reward function to guide training in each time step. However, in real world, it is non-trivial to design such a reward function, and the only signal available is usually obtained at the end of a trajectory, also known as the episodic reward or return. In this work, we introduce a new algorithm for temporal credit assignment, which learns to decompose the episodic return back to each time-step in the trajectory using deep neural networks. With this learned reward signal, the learning efficiency can be substantially improved for episodic reinforcement learning. In particular, we find that expressive language models such as the Transformer can be adopted for learning the importance and the dependency of states in the trajectory, therefore providing high-quality and interpretable learned reward signals. We have performed extensive experiments on a set of MuJoCo continuous locomotive control tasks with only episodic returns and demonstrated the effectiveness of our algorithm.
\end{abstract}

\section{Introduction}
Deep reinforcement learning (RL) methods, including the well-known policy gradient algorithms \citep{mnih2016asynchronous, schulman2015high, schulman2017proximal} 
and deep Q-networks \citep{mnih2015human}, have shown superior performance and great potential in many difficult real-world problems, such as the Go game 
\citep{silver2016mastering, silver2017mastering2}, locomotive continuous control problems \cite{lillicrap2015continuous}, resource management \citep{mao2016resource}, and
robotics \cite{levine2016end}. The key idea of such algorithms is to use deep neural networks as functional approximators to abstract or represent complex state observation so that actions can be properly chosen accordingly to optimize a long-term expected return. The learned policy or Q function essentially captures the temporal structure of the sequential decision problem and decompose it to a supervised learning problem, guided by the reward signal. However, in many real-world problems, the reward signal is usually not dense enough to provide sufficient supervision for learning the decision at each single time step. In many practical tasks, such as the Go game and the automatic chemical design problems \citep{olivecrona2017molecular}, we can only obtain a final reward or return value after finishing the entire rollout of the policy, while no intermediate reward is provided before reaching the end of the trajectory. This type of problems is also known as the episodic reinforcement learning. 


Unfortunately,  when the reward signal becomes delayed or \edit{even} episodic, most existing deep reinforcement learning algorithms may get stuck during the training process and often suffer from inferior performance and inefficient sample complexity~\cite{Gangwani2018LearningSD,guo2018generative}. This problem is widely known as the temporal credit assignment in reinforcement learning \citep{Sutton:1984:TCA:911176}, which describes the issue of delayed rewards causing the signal to be diluted over time and only weakly affecting the states temporally distant from the time step when the rewards get collected. For example, in Go games, the only \edit{effective reward} is the final win or loss. This reward is received only after finishing the entire games, usually consisting of hundreds of moves in the trajectory. During the game, though human professional players or experts may be able to decide which moves are likely to influence the final winning probability, quantitatively design of such informative rewards is very challenging. Such sparse or episodic reward signal makes the training of policies or Q-neural networks very difficult, as a lot of data is required to propagate the final win/loss reward back to earlier states or state-action pairs. In addition, since there is no immediate reward or even no short-term reward, the exploration becomes challenging without any information. Similar to Go games, there are many real-world problems in which rewards are terribly delayed or episodic, with only a non-zero reward value received at the end of the episode or the trajectory. 

In this paper, we introduce a new algorithm for the temporal credit assignment problem. The idea is to learn deep neural networks that is able to decompose the episodic reward into parts and assign them back to each time step in the trajectory. After learning this decomposition, we can use the assigned dense reward signal to guide \edit{policy optimization}. Formally, we derive a generalized policy gradient for decomposed reward signal. In our derivation, we find that to ensure the correctness of the policy gradient algorithms, the dependency of the neural network on time steps needs to be forward-looking, i.e. for a time $t$ before the terminal step $T$, the reward function $r_t(\{s_{t'},{a_t'}\}_{t'=0}^T) = r_t(\{s_{t'},a_{t'}\}_{t'=0}^t)$ does not depend on any future states or actions. This forward structure, frequently seen in natural language processing~\citep{Jurafsky:2009:SLP:1214993}, motivates us to apply the renowned neural-network \edit{sequence models, such as} language models to learning the reward decomposition. In particular, we adopt the Transformer, which satisfies the forward-looking structure, for learning the importance and the dependency of states using the self-attention mechanism. With the learned reward signal, we show on a set of MuJoCo continuous locomotive control tasks with episodic returns that the learning and sample efficiency of can be greatly improved. 

\section{Method}

\subsection{Background and Problem Statement}
\paragraph{\bf Reinforcement learning} Reinforcement learning considers the problem of finding an optimal policy for an agent which interacts with an environment and collects reward per action. The goal of the agent is to maximize the cumulative reward along a trajectory. 
Formally, this problem can be formulated as a Markov decision process (MDP) over the environment states $s \in {\it S}$ and agent actions $a \in {\it A}$, 
under an unknown environmental dynamics defined by a transition probability $T(s'|s,a)$. \edit{The agent's action $a$ is selected by a conditional probability distribution $\pi_\theta(a|s)$ parameterized by $\theta \in \Theta$. Playing the policy repeatedly under MDP yields a trajectory 
$\tau = \{s_t, a_t\}_{t=0}^T$, where $T$ denotes the horizon length. 
Each trajectory $\tau$ is associated with a reward function $R(\tau)$ which we want to optimize, that is, $\max_{\theta} J(\theta):= \E_{\pi_\theta}\left [R(\tau)\right ],$
where the expectation $\E_{\pi_\theta}$ is with the distribution of the unknown dynamics $T(s_{t+1}|s_t,a_t)$ and policy $\pi_\theta(a_t|s_t)$.}
%

 In standard RL settings, it is common to assume that the reward is a discounted sum of a set of local reward functions distributed across the time, that is,  $R(\tau) = \sum_{t=0}^\infty \gamma^t r_t\left(s_t, a_t\right),$
 where the local reward signal $r(s,a)$ is assumed to be observed immediately following the action $a$ performed at state $s$, and  $\gamma\in [0,1)$ is the discount factor. This decomposition structure greatly simplifies the problem, and forms the basic assumption of popular policy gradient algorithms. The focus of this work, however, is the more difficult case when $R(\tau)$ does not have a simple or known decomposition structure priori, and is observed only after the whole trajectory $\tau$ is rolled out. 
\newcommand{\pit}{\pi}

{\bf Policy Gradient} There are several different types of algorithms for learning policies, including Q-learning, policy gradient and evolutionary algorithms. For the general episodic reward function $R(\tau)$, we have a basic gradient estimation derived using the likelihood ratio trick,  
\begin{align}\label{equ:reinforce}
\begin{split}
    \nabla_\theta J(\theta) 
& = \nabla_\theta \E_{\pi_\theta}[R(\tau)] =  \E_{\pi_\theta} \left [ R(\tau) \sum_{t=0}^T \nabla_{\theta} \log  \pi(a_t|s_t) \right ]. 
\end{split}
\end{align}

This basic algorithm, however, often yields large variance in gradient estimation. 
The policy gradient theorem \citep{sutton2018reinforcement} 
allows us to derive a simplified formula for the case when the reward is decomposed step-wisely:
\begin{equation}\label{equ:pg}
\nabla_\theta J(\theta) = 
\E_{\pi}\left[\nabla_\theta \log \pi(a | s) Q^{\pit}(s,a)\right], 
\end{equation}
where $Q^{\pit} (s,a) =\E_{\pi}\left[\sum_{t=0}^\infty \gamma^{t} r(s_t, a_t)|s_0=s, a_0=a\right]$ denotes the expected return under policy $\pit$ starting from state $s$ and action $a$.
\edit{With empirical estimate of $\hat{Q}^{\pit}(s_t, a_t) = \sum_{j \geq t} \gamma^{j - t} r_j$ using rollout trajectories, then we can obtain the well-known REINFORCE policy gradient \citep{williams1992simple} as}
\begin{equation}
\hat \nabla_\theta J(\theta) =
\frac{1}{T}\sum_{t=0}^T \gamma^{t} \nabla_\theta \log \pi(a_t | s_t) \hat Q^{\pit}(s_t, a_t).  
\end{equation}

\edit{Improved policy gradient methods, such as the Proximal Policy Optimization (PPO) \citep{schulman2017proximal, heess2017emergence}, are now able to provide the state-the-art performance on many problems. It uses a 
proximal Kullback-Leibler (KL) divergence penalty to regularize and stabilize the policy gradient. Furthermore, control variate methods, such as the generalized advantage estimation (GAE)~\cite{schulman2015high}, help reducing the variance of the policy gradient estimation. So far, most policy gradient methods are designated for infinite-horizon, dense reward settings, as the dense rewards can provide direct supervision for value function estimation and policy improvement in each time step. }

\replace{
{\bf Episodic RL} For tasks with episodic rewards, usually, there is a set of terminal states. At the end of each trajectory $\tau$, a reward is only received at the terminal state. In other words, before reaching the final state $s_T$, rewards $r_t(s_t,a_t) = 0$ for all $t<T$. In many tasks, the terminal states can be predefined, or the length of the trajectories is limited. For  simplicity, we omit the discount factor and assume the trajectory length is at most $T$. In this way, we can abuse the notation of $s_T$ to denote the last state without further confusion. Therefore, the objective of episodic reinforcement learning becomes $J(\theta) 
= \E_{\pi_\theta} \left[ R(\tau) \right] = \E_{\pi_\theta} \left[r_T(s_T)\right].$ Note that it is not hard to add the discount factor back and/or adopt the mathematical definitions and derivations for problems with a set of terminal states.}{
{\bf Episodic RL} For tasks with episodic rewards, usually, there is a set of terminal states. At the end of each trajectory $\tau$, the final state $s_T$ is a terminal state, and a reward is only received at this state. In other words, before reaching the final state $s_T$, all rewards $r_t(s_t,a_t) = 0$ for all $t<T$. In many tasks, the terminal states can be predefined, or the length of the trajectories is limited. For  simplicity, we omit the discount factor and assume the trajectory length is at most $T$. In this way, we can abuse the notation of $s_T$ to denote the last state in any trajectory, without further confusion. Therefore, the objective of episodic reinforcement learning becomes $J(\theta) 
= \E_{\pi_\theta} \left[ R(\tau) \right] = \E_{\pi_\theta} \left[r_T(s_T)\right].$ Note that it is not hard to add the discount factor back and/or adopt the mathematical definitions and derivations for problems with a set of terminal states.}

For episodic problems, the straightforward application of policy gradient methods, including REINFORCE~\citep{williams1992simple}, A2C~\citep{mnih2016asynchronous} and PPO~\citep{schulman2017proximal}, may suffer from sample inefficiency, as the final episodic reward would only provide the same or similar supervision for learning policy over all time steps in a trajectory. Therefore, a huge volume of rollout trajectories are required to distinguish the subtle influence of certain action on the final reward. \replace{Besides policy optimization methods, blackbox optimization approaches , such as cross entropy method~\citep{rubinstein1999cross}, CMA-ES~\citep{hansen2001completely} and evolution strategies~\citep{salimans2017evolution}, have been also applied to episodic RL problems due to their computational efficiency. }{Alternatively, blackbox optimization methods, such as cross entropy method~\citep{rubinstein1999cross}, CMA-ES~\citep{hansen2001completely} and evolution strategies~\citep{salimans2017evolution}, have been also applied to episodic RL problems.}

\subsection{Overview of Our Approach} We propose a new approach to learn a dense surrogate reward function that approximates the temporal credit assignment of the episodic reward (Algorithm~\ref{alg:algorithm}). The idea is intuitive: we hope to find $\hat{r}(s_t,a_t)$ approximating $R(\tau) = \sum_{t=0}^T \hat{r}(s_t,a_t)$, so that we can use $\hat{r}$ as the surrogate reward to compute the policy gradient. If $\hat{r}$ is dense over time step and provide sufficient information about the influence on the episodic reward, then it should help improve the sample efficiency of training policies under the episodic reward setting.

%
\begin{algorithm}[H]
    \caption{Policy optimization with decomposed reward}
    \label{alg:algorithm} 
    \begin{algorithmic}[1]
    \STATE \textbf{Initialize: policy parameters $\theta_0$, predictor parameters $\phi_0$}
    \FOR {$i$ = 1, 2, 3, ... $N$}
        \STATE Collect a batch of trajectories using roll-outs.
            \STATE Append the new trajectories into trajectory buffer for regression.
            \STATE Train reward predictor using gradient descent: $\phi_{i} \leftarrow \phi_{i -1} - \gamma_{\phi}\nabla_\phi{\mathcal{L}_{regression}}$
            \STATE Update policy parameters using policy optimization algorithm: $\theta_i \leftarrow \theta_{i-1} + \gamma_{\theta}\nabla J(\theta)$ where $\nabla J(\theta)$ is obtained with Eq.~\eqref{equ:hatpg2} .
        \ENDFOR
        \STATE \textbf{Output: policy $\pi_{\theta_N}$, reward predictor $\phi_N$}
    \end{algorithmic}
\end{algorithm}

Here, we consider a generalization of the time step-wise reward function. Assume a reward function $\hat{r}$ that is defined on states and actions over a time interval $\alpha \in \mathcal{I}$. Then we expect that the episodic reward $R(\tau) = r_T(s_T)$ can be decomposed as the sum of the reward function on all intervals: $\sum_{\alpha \in \mathcal{I}} \hat{r}(s_{\alpha}, a_{\alpha})$, where $s_\alpha = \{s_i | i\in \alpha\}$ and $a_\alpha = \{a_i | i\in \alpha\}$. The choice of $\mathcal{I}$ can be very flexible: If each interval only contains a single time step, then the reward function is defined on each time step $\alpha \in \{\{0\},\{1\},\ldots, \{T\}\}$; If $\mathcal{I}$ contains all consecutive sub-sequences starting from time $0$, then each $\alpha \in \{\{0,1,2,\ldots,t\}: \forall t=0,\ldots, T\}$ contains all time steps from the beginning of the trajectory to the current time step $t$. 

Therefore, the objective function of learning such reward $\hat{r}$ can be done by minimizing the regression loss as following:
\begin{equation}
\hspace{-0.2cm}\label{equ:regression}
    \min_{\phi} \mathcal{L}_{regression}(\phi) :=  \sum_{\tau \in D} (\sum_{\alpha \in \mathcal{I}} \hat{r}_\phi(s_{\alpha}, a_{\alpha}) - R(\tau) )^2
\end{equation}
where
$\hat{r}_\phi$ can be a neural network that takes $s_\alpha$ and $a_\alpha$ as input and is parameterized by $\phi$, $D$ is a collection of trajectories. There are several critical choices: first, how one should define the interval set $\mathcal{I}$; second, which model would be used for the reward function; third, how to collect the dataset $D$. We will discuss the design principles in the following sections and their impact in the experimental section.

\subsection{Generalized Policy Gradient with Rewards on Time Intervals}

\paragraph{Policy Gradient of Composite Reward}
Now, let us assume we have learned a composite approximation of the reward  function, $\hat R(\tau)=\sum_{\alpha \in \mathcal I} \hat{r}(s_{\alpha}, a_{\alpha})$, where $\hat r(s_\alpha, a_\alpha)$ is a local reward function that defined on state and actions over time interval $\alpha \in \mathcal I$. \replace{}{Denote by $r_0(\tau) = R(\tau) - \hat R(\tau)$ the residual error between true reward $R(\tau)$ and the estimated reward $\hat R(\tau)$.} Our key idea is to leverage the decomposition structure of $\hat R$ to simplify and reduce the variance of the policy gradient formula,
as we summarize in the following generalization of policy gradient theorem to the composite rewards.  
\begin{thm}\label{thm:main}
I) Denote by $\hat J(\theta) := \E_{\pi_\theta}[\hat R(\tau)] $ the expectation of the composite reward $\hat R(\tau)$, we have 
\begin{align}
\label{equ:control}
\nabla \hat J(\theta) 
& = \nabla_\theta \E_{\pi_\theta}[\hat R(\tau)]  
  = \sum_{\alpha \in \mathcal I}  \E_{\pi_\theta}\left [ \hat r(s_{\alpha}, a_{\alpha})  \sum_{t\in \Gamma_\alpha} \nabla_\theta \log \pi(a_t | s_t)  \right ], 
\end{align}
where $\Gamma_\alpha = \{ t \colon t \leq \max(\alpha)\}$
and $\max(\alpha)$ denotes the maximum element of set $\alpha$; 
note that $\Gamma_\alpha$ is the set of all $t$ that  $\nabla_\theta \log \pi(a_t| s_t)$ should multiply by $\hat r(s_\alpha, t_\alpha)$.

II)  Equivalently, we have \vspace{-0.1cm}
\begin{equation}
\label{equ:pg2}
\nabla \hat J(\theta) 
 = 
 \E_{\pi_\theta}\left [ \sum_{t=0}^T Q_t(\tau) 
 \nabla_\theta \log \pi(a_t | s_t)  \right ],
\end{equation}
where $Q_t$ is a generalized $Q$-function, defined to be 
$$
Q_t(\tau) = \sum_{\alpha \in \Gamma_t^*}  \hat r (s_\alpha, a_\alpha), ~~~  \Gamma_t^* = \{\alpha \colon  \max(\alpha) \geq t \}. 
$$
Here $\Gamma_t^*$ is the set of all $\alpha$ whose upper bound $\max(\alpha)$ excess $t$.  
\end{thm}
Theorem~\ref{thm:main} allows us to leverage decomposition structure of $\hat R(\tau)$ to increase the efficiency of policy gradient.
Compared to the basic gradient formula in \eqref{equ:reinforce}, 
Eq.~\eqref{equ:control} keeps only the $\nabla \log \pi(a_t|s_t)$ in set $\Gamma_\alpha$ for each local reward $\hat r(s_\alpha, a_\alpha).$ 
This is obtained by the fact of MDP that present actions only impact the future but not the past, similar to what we have seen in the original policy gradient.

Eq.~\eqref{equ:pg2} in Theorem~\ref{thm:main} can be viewed as a generalization of policy gradient theorem in Eq.~\ref{equ:pg}, and $Q_t$ is similar to the definition of typical notion of $Q$-function, which corresponds to our special case when each $
\alpha \in \mathcal I$  includes an individual time steps, that is, $\mathcal I = \{\{0\},\{1\},\ldots, \{T\} \}$.  

By replacing the expectation in  Eq.~\eqref{equ:pg2} with empirical average from \replace{rollout trajectories}{empirical data}, we can derive a generalized policy gradient for the composite reward $\hat R(\tau),$ that is, 
given a set of trajectories $\tau^i = \{(s_t^i, a_t^i)_{t=0}^T\}$, $i =1,\ldots, n$, we have 
$$
\nabla_\theta \hat J(\theta) 
= \frac{1}{n}\sum_{i=1}^n \sum_{t=0}^T  
Q_t(\tau^i) \nabla \log \pi(a_t^i | s_t^i), 
$$
where again $
Q_t(\tau^i)  = \sum_{\alpha\in \Gamma_t^*} \hat r(s_\alpha^i, a_\alpha^i)$ is the generalized $Q$ function.
\paragraph{Bias Correction and Control Variates}
 The method above works well if $\hat R(\tau)$ forms an accurate approximation of $R(\tau)$. 
 However, when the approximation is poor, 
 it introduces a significant bias into the gradient estimation and hence deteriorates the performance.
 To address this problem, we propose to add the residual term to correct this bias. This yields a gradient estimation of form 
 \begin{align}\label{equ:hatpg2}
\nabla J(\theta) 
= \E_{\pi_\theta}\left [r_0(\tau) \sum_{t=0}^T
\nabla \log \pi(a_t|s_t) \right ] ~+~\nabla \hat J(\theta). 
 \end{align}
 where $r_0(\tau) = R(\tau) - \hat R(\tau)$ denotes the residual error, 
 and $\nabla \hat J(\theta)$ is the gradient of
 composite reward  $\hat J(\theta) = \E_{\pi_\theta}[\hat R(\tau)]$ in  \eqref{equ:pg2} and \eqref{equ:hatpg2}. This allows us to give an unbiased gradient estimation of the true expected reward $J(\theta)$, while being able to leverage the structure of the composite reward. 
 
 Theoretically, we can show that 
 the residual corrected gradient estimation \eqref{equ:hatpg2} can be viewed as a control variate, a widely used approach for reducing variance in policy optimization. 
 In particular, we can show that \eqref{equ:hatpg2} is equivalent to 
\begin{equation}
\label{equ:pg3}
\hspace{-0.25cm}\nabla J(\theta) 
 = \E_{\pi_\theta}\left [  \sum_{t=0}^T \left  (R(\tau) - \hat r_{\neg t}(\tau) \right )\nabla_\theta \log \pi( a_t | s_t) \right]
\end{equation}

where $\hat r_{\neg t}(\tau) = \sum_{\alpha \notin \Gamma_t^*} \hat r(s_\alpha, a_\alpha)$, which collects all the terms that do not appear in \eqref{equ:control}. 
We can show that $\E_{\pi_\theta}[\hat r_{\neg t}(\tau)\nabla_\theta \log \pi( a_t | s_t) ]=0$ for all $t$, and hence subtracting $r_{\neg t}$ in \eqref{equ:pg3} does not change the expectation of the formula. Note that $r_{\neg t}$ is a generalization of  the standard baseline, which is typically taken to be a constant, or depend only on $s_t$. 

With this unbiased estimator, the policy gradient would be robust so that we no longer need to worry about whether the regression of $\hat r$ is sufficient or the dataset $D$ is not well chosen. In practice, we find that both $\nabla_\theta \hat J(\theta)$ and $\nabla_\theta J(\theta)$ works reasonably well if the regression loss is sufficiently optimized.

\paragraph{Discussion.} Note that if $\mathcal I$ is the set of individual time steps and $R(\tau) = \sum_t \hat r(s_t,a_t)$, Eq.~\eqref{equ:pg3} will be reduced to the classic policy gradient as shown in Eq.~\eqref{equ:pg}. It is probably more interesting to consider $\mathcal I$ to go beyond the set of time steps to include more information. One critical observation from the above derivation is that the interval $\alpha$ can be as large as possible if $\max(\alpha)$ is upper bounded by time $t$, if we want the supervision of $\hat r(s_\alpha,a_\alpha)$ for $\nabla_\theta \log \pi_\theta(s_t,a_t)$. Therefore, to include as much information as possible, it is natural to see that we can define $\mathcal I$ as the set of all consecutive sub-sequences starting from time $0$, so that each $\alpha \in \{\{0,1,2,\ldots,t\}: \forall t=0,\ldots, T\}$ contains all time steps from the beginning of the trajectory to the current time step $t$. This structure of $\mathcal I$ is particularly interesting and highly resembles the forward-looking structure studied in sequence modeling in NLP, such as language modeling.

\subsection{Learning Temporal Credit Assignment using Language Models}
Motivated by the forward-looking structure, we propose to adopt neural-network language models for learning the reward function $\hat r$ for credit assignment. In nature language processing, a language model assigns probability to a given sequence in a language~\citep{bengio2003neural}. A more tangible and related model is to assign probability of an upcoming word given a sequence of prior words. \replace{More formally, a language model predicts the probability of word $w_t$ by parameterizing the conditional distribution $p(w_t|w_1,w_2,\ldots,w_{t-1})$.}{More formally, a language model $p(w_t|w_1,w_2,\ldots,w_{t-1})$ is able to predict the probability of word $w_t$.} Such models would be very useful in many applications, especially those generating sequences as output. Notable examples including the n-gram models and the recurrent neural networks, which attempt to capture medium- to long-range dependencies in the sentence. Very recently, a model entirely based on attention mechanisms was proposed for language modeling and has achieved state-of-the-art performance on neural machine translation and other NLP tasks ~\cite{vaswani2017attention, devlin2018bert}. The model, called Transformer, has an encoder-decoder structure and is composed of stacked self-attention and fully connected layers, without using any recurrence or convolution. To attend multiple parts of the input sequence simultaneously, instead of using a single large attention ``head", the Transformer uses multiple small attention heads to project the input sequence into multiple subspaces and combines the attention outputs by concatenation. Note that in Transformer, the self-attention for constructing an latent vector at a word is based on all other words prior to the current one, which highly resembles the desideratum for the reward function, as discussed above. Furthermore, it is also possible to use recurrent neural networks or convolutional neural networks \edit{to model} the reward function. 

In this work, we consider the Transformer network model, because of its superior performance and interpretability. Our model for learning the reward function $\hat r$ consists of two parts. The first part is an encoder module of the Transformer, which is composed of a multi-head attention layer followed by a position-wise fully-connected feed-forward layer. The second part is a self-attention layer~\citep{lin2017structured}, which generates a set of summation weight vectors for the encoder outputs. The set of summation weight vectors are used to multiply with the Transformer outputs, resulting in a hidden representation which is then processed by a regression layer to give the predicted rewards $\hat r$.

Specifically, suppose we have a trajectory that has $n$ state-action pairs, represented as $\tau=\{(s_t, a_t)^{T-1}_{t=0}\}$. Here $s_t$ is a $d_s$-dimensional observation vector and $a_t$ is a $d_a$-dimensional action vector. $\tau$ is thus represented as a $T$ by $(d_s+d_a)$ matrix. Note that we omit the special terminal state $s_T$ here for notational simplicity, if not causing further confusion. Each state-action pair $(s_t, a_t)$ in the trajectory is then processed by a feed forward layer, whose parameters are shared across all time steps, to give a fixed-length vector representation $\mathbf{v}_t$ for this state-action pair. To gain the dependency between the current time step $t$ and other time steps before $t$, we use a encoder layer to process the state-action pairs: $\mathbf{h}_t=\text{Transformer}(\mathbf{v}_0, \mathbf{v}_1, \ldots, \mathbf{v}_t)$.
Let the dimension of each start-action pair representation vector $\mathbf{v}_t$ be $d$. Since the Transformer network does not change the dimension of input vectors, we represent all the $T$ representation vectors $\mathbf{h}_t$ as a $T$ by $d$ matrix $H=(\mathbf{h}_0, \mathbf{h}_1, \ldots, \mathbf{h}_{T-1})$. To summarize the information in $H$, we apply a self-attention mechanism:
\begin{equation*}
  \mathbf{z}=\text{sigmoid}(\mathbf{w}_{s2}\tanh(W_{s1}H^{\top}))^{\top}.  
\end{equation*}
Here $W_{s1}$ is a weight matrix with dimension $d_z$ by $d$ and $\mathbf{w}_{s2}$ is a vector of parameters with size $d_z$, where $d_z$ is a hyper-parameter. Vector $\mathbf{z}$ has size $T$, and each entry $z_t$ ranges from 0 to 1, quantifying the importance of the state-action pair $(s_t, a_t)$ in predicting the reward $r(s_{\alpha_t}, a_{\alpha_t})$ for interval $\alpha_t = \{0,1,\ldots,t\}$. We combine the hidden representations in $H$ using $\mathbf{z}$ to obtain a summarized representation $\mathbf{h}^*_t=z_t\mathbf{h}_t$. To predict the reward $\hat r(s_{\alpha_t}, a_{\alpha_t})$, we add one regression layer parameterized by $\mathbf{w}_r$ and $b_r$ and output the predicted reward as 
\begin{equation*}
    \hat r(s_{\alpha_t}, a_{\alpha_t})=\mathbf{w}_r^{\top}\mathbf{h^*_t}+b_r.
\end{equation*}

Some other networks we consider include the feed-forward neural network and long short-term memory network (LSTM). Please see Figure~\ref{fig:networks} to see their differences.

\section{Experiments}
In this section, we conduct experiments to provide evidences for the following questions?
(1) Is the learned reward function useful for improving policy optimization in episodic RL?
(2) What are the appropriate choices of interval set $\mathcal{I}$, neural network model, and dataset $D$?
\edit{Note that we also provide a case study on visualization and intepretability of the learned reward function in Appendix.}

\subsection{Experimental Settings}
The experiments are conducted on a set of high-dimensional locomotion tasks in continuous domain using OpenAI gym~\citep{brockman2016openai} and MuJoCo simulation toolkits. We use the PPO algorithm introduced in \cite{schulman2017proximal} for all the experiments. The policy is represented by a uni-mode Gaussian distribution with diagonal covariance. The mean is parameterized by a two-layer neural network with 64 hidden units and \textit{tanh} non-linearity. The log standard deviation is parameterized by a global vector. The same architecture is applied for value function approximation. In practice, we implemented the policy gradient estimation shown in Eq. \eqref{equ:hatpg2}, where we decompose into two parts: one with the predicted reward, the other with only the residual. In this way, we can use advanced variance reduction approaches to improve the estimation of each components. In our experiments, we use the generalized advantage estimation (GAE) \cite{schulman2015high} as the control variate method for variance reduction. More hyper-parameters of PPO and GAE are tabulated in Table~\ref{tab:hyper_parameters} in Appendix. We also compare against baseline algorithms trained with episodic reward, respectively. The episodic reward of each rollout trajectory is defined as the accumulated original reward at all time-steps. The experiments are run for 5M timesteps with 5 random seeds. \edit{For baseline algorithms, we also consider an LSTM policy with 128 hidden units besides the aforementioned MLP policy. All experiments are conducted on NVIDIA 1080 GPUs. }

\begin{table*}[t!]
    \centering
    \begin{tabular}{c|c|c|c|c|c}
\hline
    			 & Hopper & Walker2d & Humanoid & Humanoid-Standup & Swimmer \\ 
\hline
PPO (episodic)	&	437	    &   266     &	516	    &   44673   &   6   \\
CEM             &   97      &   205     &   426     &   $\approx 9.6\times 10^4$   &  17  \\
\hline
Ours           &   1462	&   3217	&   2209	&   82579	& 135   \\
\hline
    \end{tabular}
    \caption{Performance of PPO and our approach on locomotion tasks with episodic rewards. Scores taken at 5M iterations with the environment. Cross entropy method performance taken from \citep{Gangwani2018LearningSD}. }
    \label{tab:my_label}
    \vspace{-0.5cm}
\end{table*}

For the reward function in our experiments, we investigate three neural network architectures, \textbf{Feed-Forward (FF) Network}, \textbf{LSTM}~\citep{hochreiter1997long}, and \textbf{Transformer}, to identify the best network structure that captures temporal dependency (Figure~\ref{fig:networks}). 
For observation and action $s_t, a_t$ at time $t$, an FF network, whose parameter $\phi$ is shared across all time steps, is used to give the predicted reward $\hat r_t$ for individual time steps. 
LSTM's reward prediction is also conditioned on all previous timesteps, so is a function of trajectory segment $s_{0:t}, a_{0:t}$. 
\edit{To perform reward regression based on the LSTM/FF outputs, we aggregate the the representations using a mean-pooling layer. The hyper-parameters of LSTM and FF are also listed in \ref{tab:hyper_parameters}.}


For the buffer updating schemes, we proposed the following three approaches to renew the trajectory buffer for reward predicting. 
\vspace{-0.2cm}
\begin{itemize}
    \item \textbf{Online (O):} The buffer is implemented as a FIFO queue with length $K$ (hyper-parameter). In each iteration, the new roll-out trajectories will be inserted into the queue and the old trajectories will be removed if the queue is too long. 
    \item \textbf{Historical+Online (HO):} Two queues are maintained: One queue is the same as the one in \textbf{Online}, storing the most recent rollouts by the current policy. The other queue stores the trajectories with the highest episodic returns from previous rollouts. 
    \item \textbf{Stratified-Sampling (S):} 
    To balance the training with episodic return regression, the buffer stores a larger number of trajectories in the history. In each iteration, $K$ trajectories are sampled to ensure the episodic return is uniformly in the five bins. The queue size $L$ and sample queue size $K$ are hyper-parameters.
\end{itemize}

\subsection{Credit Assignment Enables Policy Optimization with Episodic Reward}

\begin{figure}[H]
    \centering
    \includegraphics[width = 1.0\textwidth]{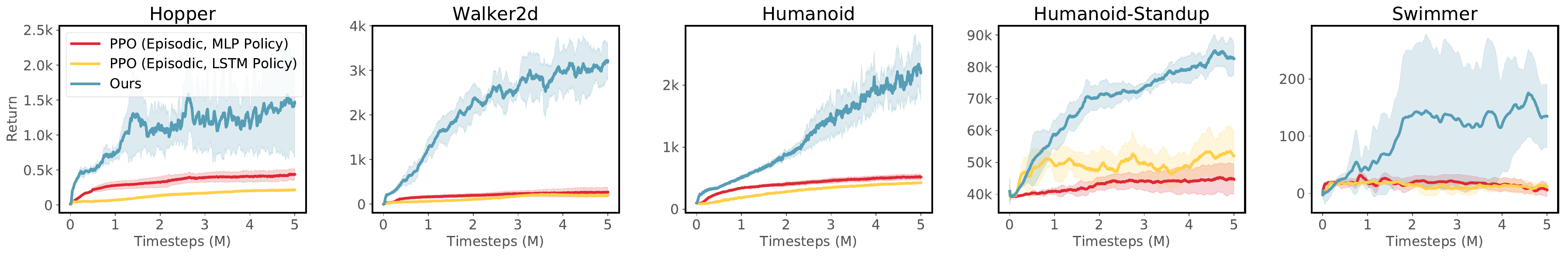}
    \vspace{-0.5cm}
    \caption{Learning curves for PPO baselines and our proposed method on tasks with episodic rewards. Mean and standard deviation over 5 random seeds is plotted. The x- and y- axis represent the number of training samples (in million) and average return, respectively.}
    \vspace{-0.5cm}
    \label{fig:exp_performance}
\end{figure}

Our experiments with the baseline algorithms on the MuJoCo control suite demonstrated that learning with episodic return is extremely hard \edit{both with MLP policy and LSTM policy}. In all the environments we tested (Figure~\ref{fig:exp_performance}), it is not surprising that the baseline method PPO (episodic) with episodic reward performs much worse than previously reported in other papers with dense reward \citep{schulman2017proximal}. In most of the runs, the policy cannot make any improvement after the initial time steps. 
On the other hand, our proposed credit-assignment algorithm with learned return decomposition consistently achieves better performance than the episodic return baselines across all environments (Figure~\ref{fig:exp_performance}). Here we use the Transformer network \edit{for decomposing rewards and MLP for representing policy,} and the HO strategy to collect trajectories for optimization. In many experiments, the policies learned by our method are able to achieve the quite reasonably good performance when the original dense reward is used for training. In environments such as the Humanoid, the Hopper and the Swimmer, with the appropriate hyper-parameters, our methods can obtain comparable performance to the policies trained with the original dense rewards, outperforming the episodic return baseline by a large margin. In addition, it also outperforms the cross-entropy method (CEM) which is suitable for the episodic RL setting (Table~\ref{tab:my_label}). These results indicate that the our reward decomposition framework can successfully enable stable learning and greatly improve the sample efficiency in episodic settings.

\subsection{Ablation Analysis}
\vspace{-0.3cm}

\begin{figure}[H]
    \centering
    \vspace{-0.4cm}
    \includegraphics[width = 1.0\textwidth]{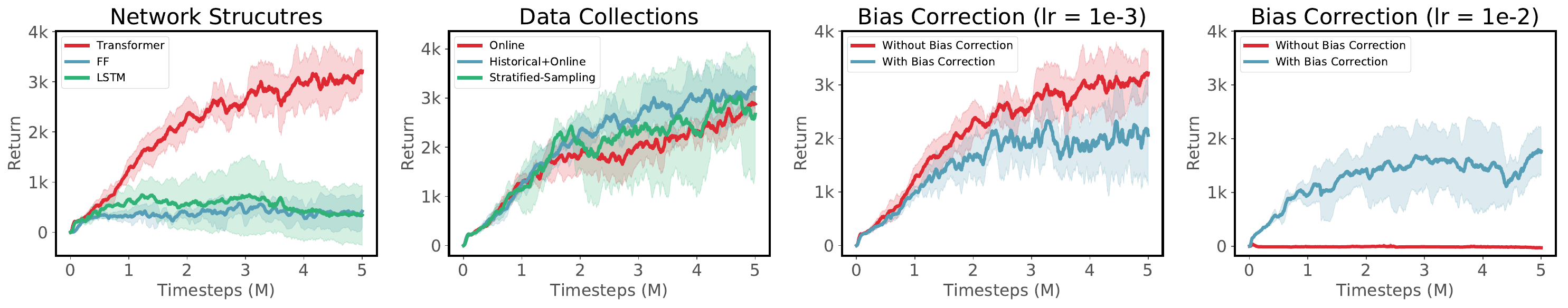}
    \vspace{-0.5cm}
    \caption{Ablation analysis on choices of (a) network structures for reward function; (b) strategies for data collection; (c)-(d) methods with or without bias correction under different learning rates (lr). Environment Walker2d was used to perform the analyses. Mean and standard deviation over 5 random seeds is plotted.}
    \vspace{-0.65cm}
    \label{fig:exp_ablation}
\end{figure}

Here we perform comparisons to check the choices of 1) three network structures for reward function; 2) strategies for data collection. We also check the utility of bias correction when the reward regression is not well fitted. We use the environment Walker2d to perform these analyses and demonstrate the results in Figure~\ref{fig:exp_ablation}.

As we can see, all the three networks structures outperform the episodic baselines, and the Transformer network provides the best performance. We conjecture that the reason why Transformer performs better than LSTM is that it is easier to train, as also implied by existing NLP literature \citep{vaswani2017attention}. Then we check whether the data collection strategy has any impact on the performance. All three strategies seem to work reasonably well, while HO appears to be the best which indicating that the historical high-quality trajectories helps reward learning. 
Finally, we compare the performance of $\nabla_\theta J(\theta)$ and $\nabla_\theta \hat J(\theta)$ under different learning rates ($10^{-2}$ and $10^{-3}$) for the regression loss. We use different learning rates essentially to adjust the quality of the reward fitting. It is easy to see that with the bias correction, the learning of $\nabla_\theta J(\theta)$ is more robust, indicating that bias correction is needed. Extra results on ablation analysis can be found in the appendix (Figures~\ref{fig:buffer_update_method}, \ref{fig:network_structure}). 


\section{Related Work}
\vspace{-0.3cm}
The optimal reward design problem \citep{singh2009rewards, sorg2010reward} concerns about finding the proxy reward function can obtains high expected return according to the true reward function. Inverse reward design \citep{hadfield2017inverse} studies the opposite problem of inferring true reward from designed reward.
Intrinsic motivation has been shown to improve sample efficiency of RL algorithms, for example by using information gain \citep{houthooft2016vime}, pseudo count \citep{bellemare2016unifying} or prediction error \citep{stadie2015incentivizing, pathak2017curiosity} as an intrinsic bonus reward to aid exploration. \citep{zheng2018learning} studies intrinsic motivation under the optimal reward design framework where the optimal intrinsic reward is learned through gradient descent. 
Reward shaping \citep{ng1999policy} explores the space of reward function modifications (specifically potential-based rewards) which do not change the corresponding optimal policy. Hindsight Experience Replay \cite{andrychowicz2017hindsight} adds additional goals and corresponding rewards to a Q-learning algorithm. Meta-learning can also be utilized to learn different objective functions for different goals \citep{houthooft2018evolved}. Our approach is more general as we don't assume specific goals of the agent. When only expert demonstrations are available as in imitation learning, inverse reinforcement learning can be used to recover the reward function from expert trajectories \citep{abbeel2004apprenticeship}. \edit{As an alternative approach for solving sparse reward problems, the auxiliary-task approaches \cite{Jaderberg2017ReinforcementLW} focus on improving the representation by adding extra self-supervised losses. However, our proposed method learns to decompose the episodic return as reward for policy optimization directly.}

\edit{ Recently, several concurrent works have explored the same direction of decomposing the episodic return. Sparse Attentive Backtracking\cite{Ke:2018:SAB:3327757.3327863} applies the attentive mechanism for network straining, while ours focuses on learning the decomposition of the reward signal. Temporal Value Transport (TVT)~\cite{hung2018optimizing} relies on a memory reconstitutive module to retrieve past data for the visual RL problems, however, ours aims at decomposing the episodic return for general problems. As one most related work, RUDDER~\cite{arjona2018rudder} could be viewed as a special case of our work in which consecutive time-steps were used. On the other hand, our work provides a general framework with interval rewards which justifies the applicability of language models and the correctness of generalized policy gradient beyond the simple Markovian assumption. }




 


\vspace{-0.3cm}
\section{Conclusion}
\vspace{-0.3cm}
We presented a new algorithm for learning temporal credit assignment, which uses deep neural network (Transformer) to decompose the episodic reward back to each time step in the trajectory. The assigned dense reward signals obtained from the decomposition are then used to guide training algorithms of policies. We demonstrated that our credit assignment algorithm substantially improved the learning and sample efficiency on a set of MuJoCo continuous locomotive control tasks. The reward function learned by our algorithm can also be interpreted by an attention mechanism, potentially providing insights on identifying key state-action pairs that contribute to successful reinforcement learning.
\bibliographystyle{plain}

\begin{thebibliography}{10}

\bibitem{abbeel2004apprenticeship}
Pieter Abbeel and Andrew~Y Ng.
\newblock Apprenticeship learning via inverse reinforcement learning.
\newblock In {\em Proceedings of the twenty-first international conference on
  Machine learning}, page~1. ACM, 2004.

\bibitem{andrychowicz2017hindsight}
Marcin Andrychowicz, Filip Wolski, Alex Ray, Jonas Schneider, Rachel Fong,
  Peter Welinder, Bob McGrew, Josh Tobin, OpenAI~Pieter Abbeel, and Wojciech
  Zaremba.
\newblock Hindsight experience replay.
\newblock In {\em Advances in Neural Information Processing Systems}, pages
  5048--5058, 2017.

\bibitem{arjona2018rudder}
Jose~A Arjona-Medina, Michael Gillhofer, Michael Widrich, Thomas Unterthiner,
  and Sepp Hochreiter.
\newblock Rudder: Return decomposition for delayed rewards.
\newblock {\em arXiv preprint arXiv:1806.07857}, 2018.

\bibitem{bellemare2016unifying}
Marc Bellemare, Sriram Srinivasan, Georg Ostrovski, Tom Schaul, David Saxton,
  and Remi Munos.
\newblock Unifying count-based exploration and intrinsic motivation.
\newblock In {\em Advances in Neural Information Processing Systems}, pages
  1471--1479, 2016.

\bibitem{bengio2003neural}
Yoshua Bengio, R{\'e}jean Ducharme, Pascal Vincent, and Christian Jauvin.
\newblock A neural probabilistic language model.
\newblock {\em Journal of machine learning research}, 3(Feb):1137--1155, 2003.

\bibitem{brockman2016openai}
Greg Brockman, Vicki Cheung, Ludwig Pettersson, Jonas Schneider, John Schulman,
  Jie Tang, and Wojciech Zaremba.
\newblock Openai gym.
\newblock {\em arXiv preprint arXiv:1606.01540}, 2016.

\bibitem{devlin2018bert}
Jacob Devlin, Ming-Wei Chang, Kenton Lee, and Kristina Toutanova.
\newblock Bert: Pre-training of deep bidirectional transformers for language
  understanding.
\newblock {\em arXiv preprint arXiv:1810.04805}, 2018.

\bibitem{Gangwani2018LearningSD}
Tanmay Gangwani, Qiang Liu, and Jian Peng.
\newblock Learning self-imitating diverse policies.
\newblock {\em Proceedings of the International Conference on Learning
  Representations (ICLR)}, 2019.

\bibitem{guo2018generative}
Yijie Guo, Junhyuk Oh, Satinder Singh, and Honglak Lee.
\newblock Generative adversarial self-imitation learning.
\newblock {\em arXiv preprint arXiv:1812.00950}, 2018.

\bibitem{hadfield2017inverse}
Dylan Hadfield-Menell, Smitha Milli, Pieter Abbeel, Stuart~J Russell, and Anca
  Dragan.
\newblock Inverse reward design.
\newblock In {\em Advances in Neural Information Processing Systems}, pages
  6765--6774, 2017.

\bibitem{hansen2001completely}
Nikolaus Hansen and Andreas Ostermeier.
\newblock Completely derandomized self-adaptation in evolution strategies.
\newblock {\em Evolutionary computation}, 9(2):159--195, 2001.

\bibitem{heess2017emergence}
Nicolas Heess, Srinivasan Sriram, Jay Lemmon, Josh Merel, Greg Wayne, Yuval
  Tassa, Tom Erez, Ziyu Wang, Ali Eslami, Martin Riedmiller, et~al.
\newblock Emergence of locomotion behaviours in rich environments.
\newblock {\em arXiv preprint arXiv:1707.02286}, 2017.

\bibitem{hochreiter1997long}
Sepp Hochreiter and J{\"u}rgen Schmidhuber.
\newblock Long short-term memory.
\newblock {\em Neural computation}, 9(8):1735--1780, 1997.

\bibitem{houthooft2018evolved}
Rein Houthooft, Richard~Y Chen, Phillip Isola, Bradly~C Stadie, Filip Wolski,
  Jonathan Ho, and Pieter Abbeel.
\newblock Evolved policy gradients.
\newblock {\em arXiv preprint arXiv:1802.04821}, 2018.

\bibitem{houthooft2016vime}
Rein Houthooft, Xi~Chen, Yan Duan, John Schulman, Filip De~Turck, and Pieter
  Abbeel.
\newblock Vime: Variational information maximizing exploration.
\newblock In {\em Advances in Neural Information Processing Systems}, pages
  1109--1117, 2016.

\bibitem{hung2018optimizing}
Chia-Chun Hung, Timothy Lillicrap, Josh Abramson, Yan Wu, Mehdi Mirza, Federico
  Carnevale, Arun Ahuja, and Greg Wayne.
\newblock Optimizing agent behavior over long time scales by transporting
  value.
\newblock {\em arXiv preprint arXiv:1810.06721}, 2018.

\bibitem{Jaderberg2017ReinforcementLW}
Max Jaderberg, Volodymyr Mnih, Wojciech Czarnecki, Tom Schaul, Joel~Z. Leibo,
  David Silver, and Koray Kavukcuoglu.
\newblock Reinforcement learning with unsupervised auxiliary tasks.
\newblock {\em CoRR}, abs/1611.05397, 2017.

\bibitem{Jurafsky:2009:SLP:1214993}
Daniel Jurafsky and James~H. Martin.
\newblock {\em Speech and Language Processing (2Nd Edition)}.
\newblock Prentice-Hall, Inc., Upper Saddle River, NJ, USA, 2009.

\bibitem{Ke:2018:SAB:3327757.3327863}
Nan~Rosemary Ke, Anirudh Goyal, Olexa Bilaniuk, Jonathan Binas, Michael~C.
  Mozer, Chris Pal, and Yoshua Bengio.
\newblock Sparse attentive backtracking: Temporal credit assignment through
  reminding.
\newblock In {\em Proceedings of the 32Nd International Conference on Neural
  Information Processing Systems}, NIPS'18, pages 7651--7662, USA, 2018. Curran
  Associates Inc.

\bibitem{levine2016end}
Sergey Levine, Chelsea Finn, Trevor Darrell, and Pieter Abbeel.
\newblock End-to-end training of deep visuomotor policies.
\newblock {\em The Journal of Machine Learning Research}, 17(1):1334--1373,
  2016.

\bibitem{lillicrap2015continuous}
Timothy~P Lillicrap, Jonathan~J Hunt, Alexander Pritzel, Nicolas Heess, Tom
  Erez, Yuval Tassa, David Silver, and Daan Wierstra.
\newblock Continuous control with deep reinforcement learning.
\newblock {\em ICLR}, 2016.

\bibitem{lin2017structured}
Zhouhan Lin, Minwei Feng, Cicero Nogueira~dos Santos, Mo~Yu, Bing Xiang, Bowen
  Zhou, and Yoshua Bengio.
\newblock A structured self-attentive sentence embedding.
\newblock {\em arXiv preprint arXiv:1703.03130}, 2017.

\bibitem{mao2016resource}
Hongzi Mao, Mohammad Alizadeh, Ishai Menache, and Srikanth Kandula.
\newblock Resource management with deep reinforcement learning.
\newblock In {\em Proceedings of the 15th ACM Workshop on Hot Topics in
  Networks}, pages 50--56. ACM, 2016.

\bibitem{mnih2016asynchronous}
Volodymyr Mnih, Adria~Puigdomenech Badia, Mehdi Mirza, Alex Graves, Timothy
  Lillicrap, Tim Harley, David Silver, and Koray Kavukcuoglu.
\newblock Asynchronous methods for deep reinforcement learning.
\newblock In {\em International conference on machine learning}, pages
  1928--1937, 2016.

\bibitem{mnih2015human}
Volodymyr Mnih, Koray Kavukcuoglu, David Silver, Andrei~A Rusu, Joel Veness,
  Marc~G Bellemare, Alex Graves, Martin Riedmiller, Andreas~K Fidjeland, Georg
  Ostrovski, et~al.
\newblock Human-level control through deep reinforcement learning.
\newblock {\em Nature}, 518(7540):529, 2015.

\bibitem{ng1999policy}
Andrew~Y Ng, Daishi Harada, and Stuart Russell.
\newblock Policy invariance under reward transformations: Theory and
  application to reward shaping.
\newblock In {\em ICML}, volume~99, pages 278--287, 1999.

\bibitem{olivecrona2017molecular}
Marcus Olivecrona, Thomas Blaschke, Ola Engkvist, and Hongming Chen.
\newblock Molecular de-novo design through deep reinforcement learning.
\newblock {\em Journal of cheminformatics}, 9(1):48, 2017.

\bibitem{pathak2017curiosity}
Deepak Pathak, Pulkit Agrawal, Alexei~A Efros, and Trevor Darrell.
\newblock Curiosity-driven exploration by self-supervised prediction.
\newblock In {\em International Conference on Machine Learning (ICML)}, volume
  2017, 2017.

\bibitem{rubinstein1999cross}
Reuven Rubinstein.
\newblock The cross-entropy method for combinatorial and continuous
  optimization.
\newblock {\em Methodology and computing in applied probability},
  1(2):127--190, 1999.

\bibitem{salimans2017evolution}
Tim Salimans, Jonathan Ho, Xi~Chen, Szymon Sidor, and Ilya Sutskever.
\newblock Evolution strategies as a scalable alternative to reinforcement
  learning.
\newblock {\em arXiv preprint arXiv:1703.03864}, 2017.

\bibitem{schulman2015high}
John Schulman, Philipp Moritz, Sergey Levine, Michael Jordan, and Pieter
  Abbeel.
\newblock High-dimensional continuous control using generalized advantage
  estimation.
\newblock {\em Proceedings of the International Conference on Learning
  Representations (ICLR)}, 2016.

\bibitem{schulman2017proximal}
John Schulman, Filip Wolski, Prafulla Dhariwal, Alec Radford, and Oleg Klimov.
\newblock Proximal policy optimization algorithms.
\newblock {\em arXiv preprint arXiv:1707.06347}, 2017.

\bibitem{silver2016mastering}
David Silver, Aja Huang, Chris~J Maddison, Arthur Guez, Laurent Sifre, George
  Van Den~Driessche, Julian Schrittwieser, Ioannis Antonoglou, Veda
  Panneershelvam, Marc Lanctot, et~al.
\newblock Mastering the game of go with deep neural networks and tree search.
\newblock {\em nature}, 529(7587):484, 2016.

\bibitem{silver2017mastering2}
David Silver, Julian Schrittwieser, Karen Simonyan, Ioannis Antonoglou, Aja
  Huang, Arthur Guez, Thomas Hubert, Lucas Baker, Matthew Lai, Adrian Bolton,
  et~al.
\newblock Mastering the game of go without human knowledge.
\newblock {\em Nature}, 550(7676):354, 2017.

\bibitem{singh2009rewards}
Satinder Singh, Richard~L Lewis, and Andrew~G Barto.
\newblock Where do rewards come from.
\newblock In {\em Proceedings of the annual conference of the cognitive science
  society}, pages 2601--2606, 2009.

\bibitem{sorg2010reward}
Jonathan Sorg, Richard~L Lewis, and Satinder~P Singh.
\newblock Reward design via online gradient ascent.
\newblock In {\em Advances in Neural Information Processing Systems}, pages
  2190--2198, 2010.

\bibitem{stadie2015incentivizing}
Bradly~C Stadie, Sergey Levine, and Pieter Abbeel.
\newblock Incentivizing exploration in reinforcement learning with deep
  predictive models.
\newblock {\em arXiv preprint arXiv:1507.00814}, 2015.

\bibitem{sutton2018reinforcement}
Richard~S Sutton and Andrew~G Barto.
\newblock {\em Reinforcement learning: An introduction}.
\newblock MIT press, 2018.

\bibitem{Sutton:1984:TCA:911176}
Richard~Stuart Sutton.
\newblock {\em Temporal Credit Assignment in Reinforcement Learning}.
\newblock PhD thesis, University of Massachusetts Amherst, 1984.
\newblock AAI8410337.

\bibitem{vaswani2017attention}
Ashish Vaswani, Noam Shazeer, Niki Parmar, Jakob Uszkoreit, Llion Jones,
  Aidan~N Gomez, {\L}ukasz Kaiser, and Illia Polosukhin.
\newblock Attention is all you need.
\newblock In {\em Advances in Neural Information Processing Systems}, pages
  5998--6008, 2017.

\bibitem{williams1992simple}
Ronald~J Williams.
\newblock Simple statistical gradient-following algorithms for connectionist
  reinforcement learning.
\newblock {\em Machine learning}, 8(3-4):229--256, 1992.

\bibitem{zheng2018learning}
Zeyu Zheng, Junhyuk Oh, and Satinder Singh.
\newblock On learning intrinsic rewards for policy gradient methods.
\newblock In {\em Advances in Neural Information Processing Systems}, pages
  4649--4659, 2018.

\end{thebibliography}

\appendix
\onecolumn 
\setcounter{figure}{0}
\renewcommand\thefigure{A\arabic{figure}}
\setcounter{table}{0}
\renewcommand\thetable{A\arabic{table}}

\section{Method Overview and Network structures}

\begin{figure}[htbp]
    \centering
    \includegraphics[width = 0.5\textwidth]{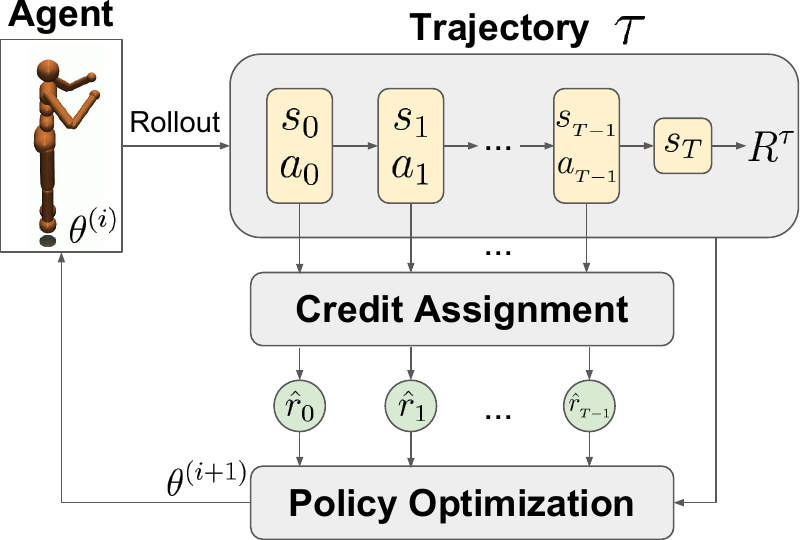}
    \caption{  Overview of our approach. Rollout trajectories are generated from interacting with the environment. A reward predictor is trained on the collected trajectories and episodic returns with the regression loss. Then the predicted rewards are used for policy optimization.}
    \label{fig:method}
\end{figure}

\begin{figure}[htbp]
    \centering
    \includegraphics[width = 0.95\textwidth]{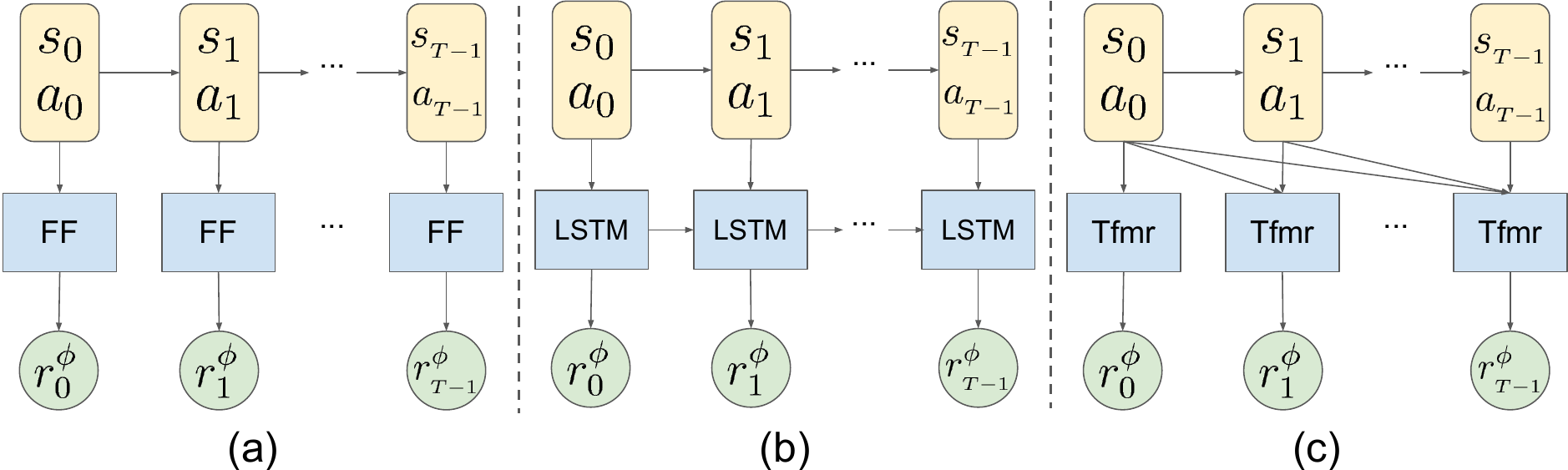}
    \caption{ Network structures for the reward predictor: (a) Feed-forward network, (b) LSTM network, (c) Transformer network }
    \label{fig:networks}
\end{figure}

\section{Proof of Theorem~\ref{thm:main}}
I) Recall the standard likelihood ratio gradient formula:
\begin{align}
    \nabla_\theta \hat J(\theta) &  = \nabla_\theta \E_{\pi_\theta}[\hat R(\tau)]  \notag \\
    & =\E_{\pi_\theta} \left [\hat R(\tau)\sum_{t=0}^T \nabla_\theta \log\pi(a_t|s_t) \right ] \notag \\
    & =\E_{\pi_\theta} \left [\left (\sum_{\alpha \in \mathcal{I}} \hat r(s_\alpha, a_\alpha)\right ) \left (\sum_{t=0}^T \nabla_\theta \log\pi(a_t|s_t) \right) \right ] \notag \\
    & =\sum_{\alpha \in \mathcal{I}} \E_{\pi_\theta} \left [  \hat r(s_\alpha, a_\alpha) \sum_{t=0}^T \nabla_\theta \log\pi(a_t|s_t)  \right ]. \label{c}
\end{align}
On the other hand, note that for any $t\notin \Gamma_\alpha$, we have 
\begin{align*}
\E_{\pi_\theta} [ \hat r(s_\alpha, s_\alpha) \nabla_\theta \log \pi(a_t | s_t)]  
= 0. 
\end{align*}
 Therefore, all the pairs $(\alpha, t)$ with $t\notin \Gamma_\alpha$ is removed in \eqref{c}. This hence yields \eqref{equ:control}. 
 
 II) Eq.~\eqref{equ:pg2} is a simple rearrangement of Eq.~\eqref{equ:control}. 
 \begin{align*}
     \nabla_\theta \hat J(\theta) &  = 
     \sum_{\alpha \in \mathcal{I}} \E_{\pi_\theta} \left [  \hat r(s_\alpha, a_\alpha) \sum_{t\in\Gamma_\alpha}  \nabla_\theta \log\pi(a_t|s_t)  \right ] \\
     & =      \E_{\pi_\theta} \left [ \sum_{t=0}^T \left (\sum_{\alpha\colon t \ni \Gamma_\alpha}   \hat r(s_\alpha, a_\alpha)  \right) \nabla_\theta \log\pi(a_t|s_t)  \right ],
 \end{align*}
 where  $(\sum_{\alpha\colon t \ni \Gamma_\alpha}   \hat r(s_\alpha, a_\alpha)  = Q_t(\tau)$, matching our definition. 
 This completes the proof.

\section{Hyper-parameters}
\begin{table}[h]
    \centering
    \begin{tabular}{c|c}
    \hline
    \hline
    Hyper-parameter                         &           Searching Values                                 \\
    \hline  
    Policy Network                          &           MLP with shape (64, 64)                          \\
    PPO Batch Size                          &           2048                                             \\
    PPO Mini-Batch Size                     &           64                                               \\
    PPO number of epoch per iteration       &           5                                                \\
    PPO learning rate                       &           0.0001                                           \\
    PPO clip range $\epsilon$               &           0.2                                              \\
    
   \hline
    GAE $\gamma$                            &           0.99                                             \\
    GAE $\lambda$                           &           0.95                                             \\
    
    \hline
    Buffer size                             &           50                                              \\ 
    Reward predictor learning rate          &           $0.001 $                                        \\
    Transformer number of heads             &           4                                               \\
    Transformer layer size                  &           64                                              \\
    Transformer hidden layer size           &           128                                             \\
    Transformer query/key size              &           32                                              \\
    \hline
    LSTM hidden size                        & 96 \\
    FF channels                           & $\left[128, 128, 128, 256\right]$ \\
    \hline
    \end{tabular}
    \caption{The hyper-parameters we used in the episodic MuJoCo environment.}
    \label{tab:hyper_parameters}
\end{table}

\section{More Results on Ablation Analysis}
\begin{figure}[H]
    \centering
    \includegraphics[width = 1.0\textwidth]{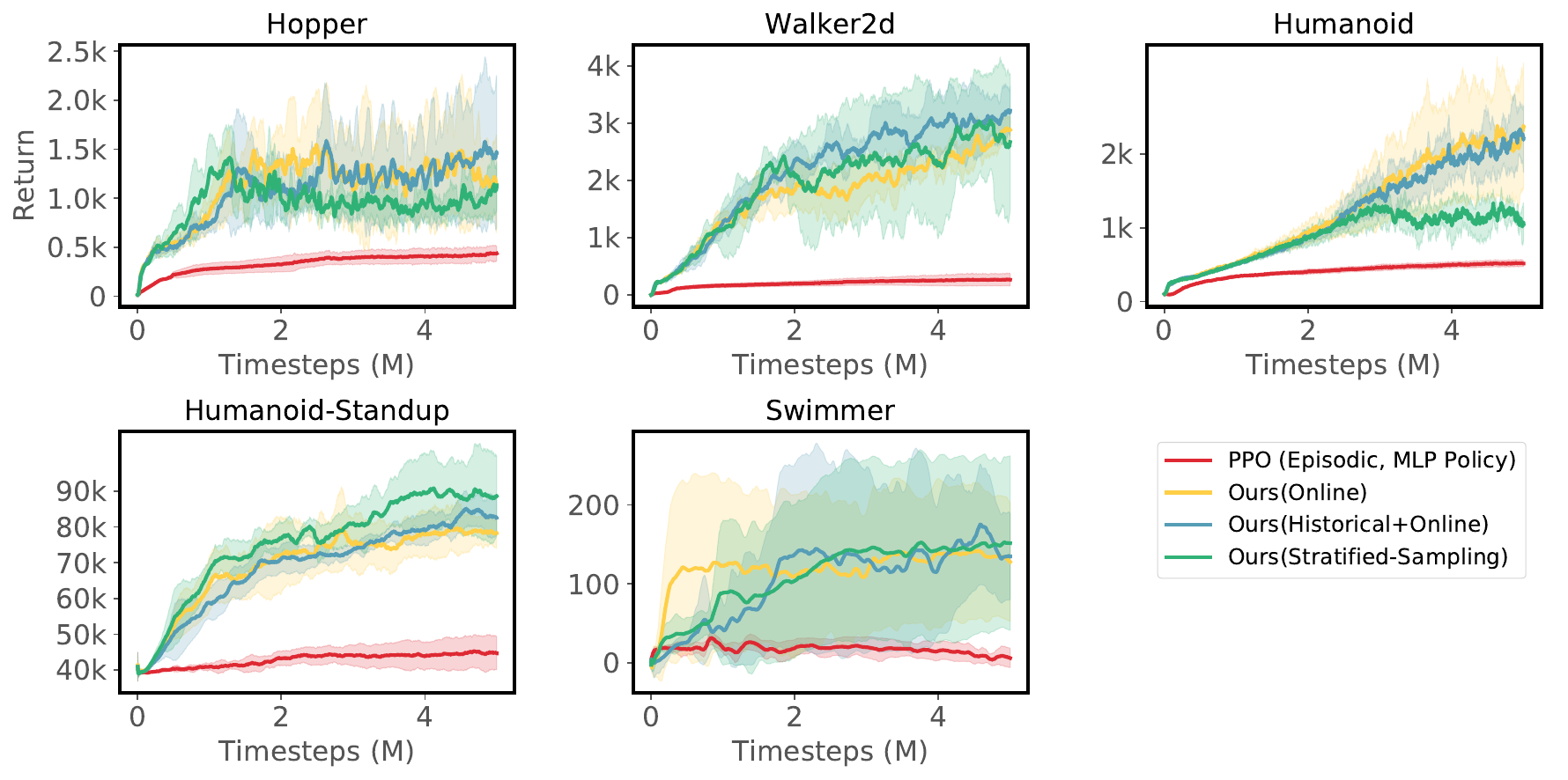}
    \caption{Comparison between different buffer updating methods. The x-axis denotes the number of training samples and y-axis denotes the average episodic return. The red curve represents the training curve using episodic return. The yellow, blue and green curves represent the algorithm with online buffer scheme, historical-online scheme and stratified-sampling scheme, respectively. }
    \label{fig:buffer_update_method}
\end{figure}

\begin{figure}[H]
    \centering
    \includegraphics[width = 1.0\textwidth]{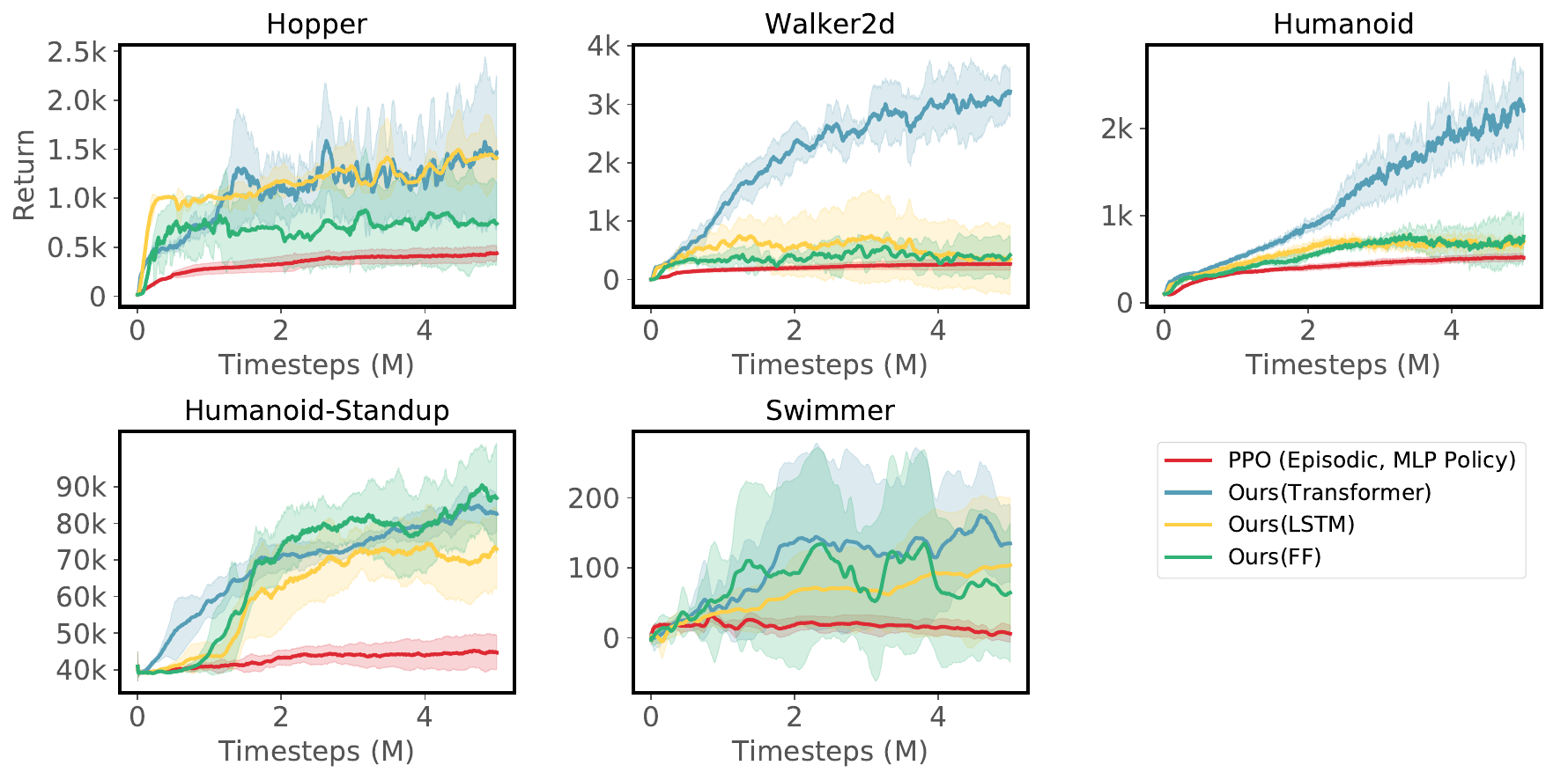}
    \caption{Comparison between network structures: The x- and y- axis represent the number of training samples and average return, respectively. The red curve represents the training curve using episodic return. The blue, yellow and green lines represent our method with a Transformer, LSTM and FF network structure, respectively. }
    \label{fig:network_structure}
\end{figure}


\section{Interpretability of the Learned Reward Function}
\begin{figure}[htbp]
    \centering
    \includegraphics[width = 0.9\textwidth]{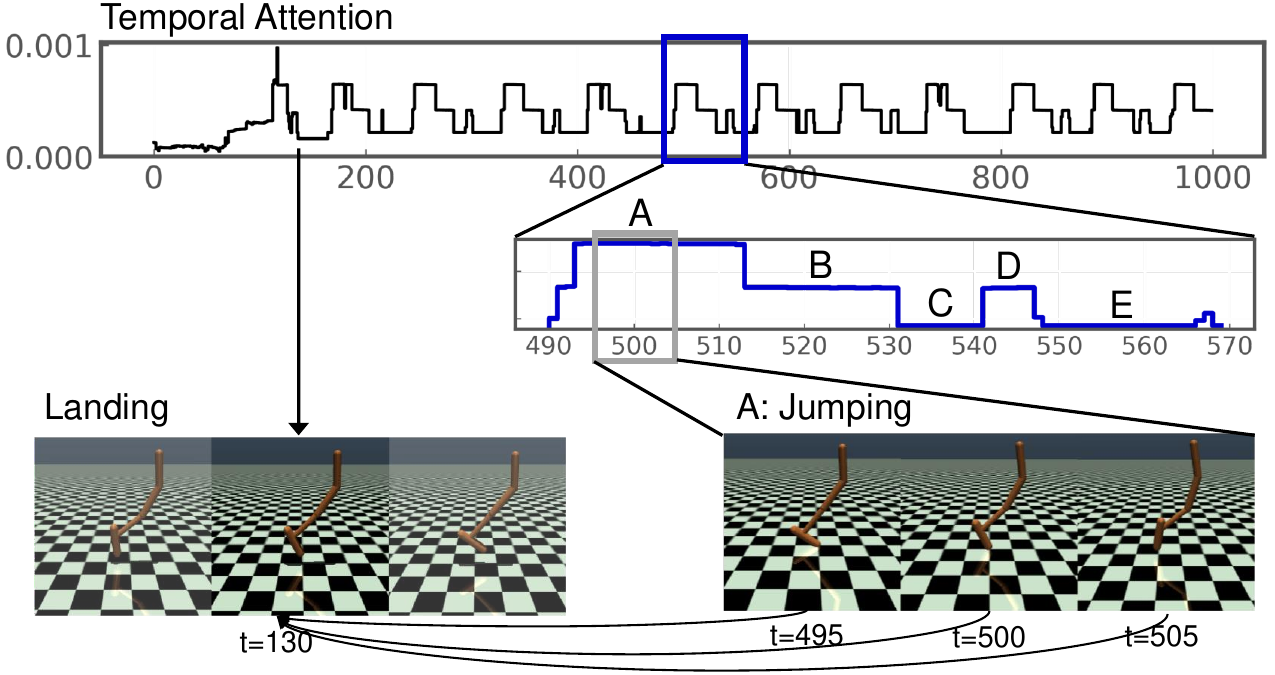}
    \caption{
    (Above) Learned temporal attention in the Transformer structure shows periodicity over time in the Hopper environment. A single period consists of five stages as marked by A to E. The model gives higher attention to jumping (A, D) than landing (B, E).
    (Below) visualizes the learned state dependency, i.e. the key reference states (e.g. landing at t=130) predicted by the multi-head attention layer in Transformer for states (e.g. jumping at t=495, 500, 505).
    We observed that the learned reward function gives higher reward and attention to the agent's jumping, as the rewards of both the jumping in the large hop (stage A) and the small hop (stage D) are relatively higher than that of landing phases (stages B, C and E). 
    }
    \label{fig:hopper_vis}
\end{figure}

We visualize the learned agent and reward function to demonstrate that the reward predictor could attain knowledge from interacting with environment. 
Key state action pairs that contribute to successful reinforcement learning can also identified from the visualization . Using the Hopper environment as an example, we visualized in Figure~\ref{fig:hopper_vis} the learned temporal attentions for 1000 time steps, extracted from the last layer of the Transformer network.  The first stage A corresponds to the agent starting a large jump, followed by the landing (stage B) and laying on the ground (stage C). The agent then makes a smaller hop in stage D and lands in stage E.
We observed that the temporal attentions exhibit periodic behavior that coincides with the agent's periodic motion. In the Hopper environment, the goal is to make the 2D one-legged robot move forward as fast as possible. A successfully trained agent essentially learns to hop forward periodically. Hence it is reasonable that the temporal attention and predicted reward show periodicity as the hopper jumps. Similar periodic behaviors of moving and adjusting balance were also be observed in other locomotion environments such as Humanoid.



\end{document}